\def\year{2023}\relax
\documentclass[letterpaper]{article} 
\usepackage{aaai21}  
\usepackage{times}  
\usepackage{helvet} 
\usepackage{courier}  
\usepackage[hyphens]{url}  
\usepackage{graphicx} 
\usepackage{amsmath}
\usepackage{amssymb}
\usepackage{gensymb}
\usepackage[ruled, vlined, linesnumbered]{algorithm2e}
\usepackage{multirow}

\def\UrlFont{\rm}  
\usepackage{natbib}  
\usepackage{caption} 
\usepackage{todonotes}

\title{Real-World Planning with PDDL+ and Beyond}
\author{
    Wiktor Piotrowski,\textsuperscript{\rm 1}
    Alexandre Perez\textsuperscript{\rm 1}
    \\
}
\affiliations{
    \textsuperscript{\rm 1} Palo Alto Research Center, CA, USA\\
    e-mail: wiktorpi@parc.com, alexandrecperez@gmail.com\\
}
\begin{document}

\maketitle

\begin{abstract}
Real-world applications of AI Planning often require a highly expressive modeling language to accurately capture important intricacies of target systems. Hybrid systems are ubiquitous in the real-world, and PDDL+ is the standardized modeling language for capturing such systems as planning domains. PDDL+ enables accurate encoding of mixed discrete-continuous system dynamics, exogenous activity, and many other interesting features exhibited in realistic scenarios. However, the uptake in usage of PDDL+ has been slow and apprehensive, largely due to a general shortage of PDDL+ planning software, and rigid limitations of the few existing planners. To overcome this chasm, we present Nyx, a novel PDDL+ planner built to emphasize lightness, simplicity, and, most importantly, adaptability. The planner is designed to be effortlessly customizable to expand its capabilities well beyond the scope of PDDL+. As a result, Nyx can be tailored to virtually any potential real-world application requiring some form of AI Planning, paving the way for wider adoption of planning methods for solving real-world problems. 
\end{abstract}

\section{Introduction}

Realistic planning problems require an expressive modeling language to accurately capture the innate intricacies of the modeled scenario. Indeed, most existing domain-independent planning languages, such a STRIPS~\cite{fikes1971strips}, ADL~\cite{pednault1989adl}, PDDL~\cite{mcdermott1998pddl}, and even PDDL2.1~\cite{fox2003pddl2}, lack features to describe commonplace elements of real-world systems. As a result, the vast majority of planning models are severely abstracted or limited in scope, often being forced to ignore aspects of the domain that, in the real world, have significant impact on the system's operations such as environmental phenomena.

PDDL+~\cite{fox2006modelling} is one of the most expressive modeling languages and was designed to model mixed discrete/continuous (hybrid) systems. PDDL+ is, arguably, the closest AI Planning has come to accurately representing realistic scenarios as planning domains. It has proved useful for capturing a wide range challenging AI domains from traffic control\cite{vallati-et-al:aaai-2016} and physics-based games\cite{piotrowski2021playing} to spacecraft operations\cite{piotrowski2018heuristics}. However, the resulting PDDL+ planning problems are notoriously difficult to solve. Planning with PDDL+ domains is undecidable and requires reasoning with a wide range of advanced domain features, including non-linear system dynamics, exogenous happenings, high branching factors, temporal activity, and more. 

Solving PDDL+ problems requires a powerful and efficient planner, capable of reasoning with the aforementioned aspects of PDDL+. However, development of PDDL+ planners has been slow and underwhelming. There are only a handful of available domain-independent PDDL+ planners, and each one is only suited for specific class of problems. 
Furthermore, most of them have a rigid and complicated code-base, very specific (often outdated) dependencies, and are generally too cumbersome to adapt for novel classes of planning models. Consequently, in lieu of AI Planning, the scientific community and potential target users turn to more accessible and flexible approaches which can be easily adapted or trained to solve emerging real-world challenges. Often, AI Planning can be particularly well-suited to efficiently tackle certain classes of problems and provide sought-after characteristics such as robustness, explainability, and transparency. However, it is overlooked as a viable approach simply because decision-makers are unaware of its potential and advantages. This issue is particularly evident for applications requiring more a expressive modeling language (recent example, well-suited for PDDL+ planning, but overlooked in favor of Learning methods: \cite{vouros2022automating}). There is a dire need for more flexible AI Planning tools which can efficiently tackle a wide range of real-world problems and which can be easily adapted to emerging classes of applications. 


We present Nyx, a novel lightweight PDDL+ planner designed with adaptability and simplicity at the forefront of development. Nyx is built to increase planning performance via easily implemented heuristics and exploiting high-fidelity settings configurations. In this paper, we describe the motivation behind developing Nyx, discuss the basic principles it is built on, and outline the path for adapting the planner to any potential scenario. 

Unfortunately, having an high-performing and capable planner is simply not enough to solve many realistic problems. The complexity of real-world systems often proves an insurmountable obstacle when attempting to model them as planning domains. Even the expressiveness of PDDL+ is bound to basic arithmetic operations, and anything beyond is either impossible to define (e.g., square root, logarithms) or requires numeric approximations (e.g., trigonometric functions). 
The second major contribution of our work is to overcome the expressiveness limits of PDDL+. Our planner, Nyx, has been designed to facilitate advanced extensions to the domain and domain definition language that span well beyond the current expressiveness of base PDDL+.

\section{Related Work}
To date, the language with, arguably, the most set of features relevant to real-world problems is PDDL+~\cite{fox2006modelling}. It was designed specifically as a planning standard for modeling hybrid systems (switched dynamical systems) governed by a set of differential equations and discrete mode switches. Formally defined as a mapping of planning constructs to Hybrid Automata~\cite{henzinger2000theory}, PDDL+ enables the modeling and solving of problems set in systems exhibiting both discrete mode switches and continuous flows. PDDL+ builds on the expressiveness of its predecessors, it encapsulates the entire set of features from PDDL2.1 (including continuous effects) and supplements that with the inclusion of timed-initial literals (TILs) from PDDL2.2. However, the biggest advancement over its predecessors is the inclusion of new constructs for defining exogenous activity in the domain.  
PDDL+ expresses exogenous activity using \textit{discrete events} and \textit{continuous processes}. Events represent the system's mode switches which instantaneously change the dynamics of the modeled system, whereas processes evolve the system over time dictated by a set of ordinary differential equations. Processes and events can be thought of as the world's version of durative and instantaneous actions, respectively. Exogenous activity is classed as `must-happen' constructs, effects are applied as soon as preconditions are satisfied. An agent interacts with processes and events indirectly to mitigate or support their effects.

The expressiveness of PDDL+ enables natural definitions of certain phenomena ubiquitous in real-world scenarios. However, the resulting planning problems are notoriously difficult to solve due to immense search spaces and complex system dynamics. Indeed, planning in hybrid domains is challenging because, apart from the state space explosion caused by discrete state variables, the continuous variables cause the reachability problem to become undecidable~\cite{alur95algorithmic}. As a result, efficient model implementation is crucial to solving PDDL+ problems, arguably more so than for any other planing domain definition language. 

Recent works highlight how PDDL+ expressiveness teamed with clever system modeling can achieve high planning performance despite the problems' complexity in a range of domains, including Urban Traffic Control~\cite{vallati-et-al:aaai-2016}, Angry Birds~\cite{piotrowski2021playing}, Chemical Batch Plant~\cite{della2010pddl+}, drilling~\cite{fox2018creating}, Minecraft~\cite{roberts2017automated}, UAV control~\cite{kiam2020ai}, and Atmospheric Re-entry~\cite{piotrowski2018heuristics}). Unfortunately, PDDL+ has been often misused. Indeed, even established hybrid domains are frequently only designed to test planner performance, instead of highlighting the importance and usefulness of PDDL+ models. Though, tackling meaningful planning problems requires development of general-purpose PDDL+ planners. It is our belief that working with domains that necessitate the use of PDDL+ will foster the development of efficient planners to raise the prominence of PDDL+ planning. 

Currently available PDDL+ planners vary in their performance and abilities to reason with different PDDL+ features. Real-world applications often require high-performing planners to handle large numbers of happenings, wide range of model features, and non-linear system dynamics. 

SMTPlan+~\cite{cashmore2016compilation} is a recent planner able to reason with all features of PDDL+ but it is only capable of dealing with polynomial non-linearity and does not scale well with the number of happenings. UPMurphi~\cite{della2009upmurphi} is a highly optimized PDDL+ planner which solves hybrid planning problems via a uniform discretization approach. UPMurphi can reason with the entire set of PDDL+ features and non-linear dynamics, though it is limited in performance by lack of heuristics. DiNo~\cite{piotrowski2016heuristic} is a PDDL+ planner which builds on the strengths of UPMurphi and extends it by implementing a domain-independent heuristic. However, DiNo's heuristic is computationally expensive and does not perform well on the Angry Birds domain, defeating its purpose. ENHSP~\cite{scala2016interval} is a versatile  PDDL+ planner equipped with various model-independent heuristics and numerous search configurations. It has undergone significant development over the past few years and can handle complex PDDL+ domains and also reason with advanced mathematical expressions. ENHSP has very minor limitations with respect to expressiveness. However, it heavily relies on multiple libraries and, thus, cannot be easily adapted to novel systems and scenarios. 

UPMurphi, DiNo, and ENHSP all adhere to the \textit{planning-via-discretization} paradigm. To date, it is the most viable approach to planning in hybrid systems. 


\section{Nyx Planner Implementation}

Lightness and accessibility are ubiquitous throughout the entire architecture and implementation of Nyx. 
The planner is written in Python, a high-level scripting language which facilitates rapid code writing and comprehension. Python is the world's most popular programming language\cite{tiobe2022,ieee2022spectrum}, and is particularly favored for AI and scientific computing applications. Python was chosen as Nyx's implementation language to make the planner more accessible and encourage various extensions and modifications. In contrast, most established planners are very specialized tools aimed at very narrow target audience. 
Their code-base is usually written using performance-focused languages (i.e., C/C++) and optimized to extract every bit of performance available. While the performance gains are significant, the software becomes much less flexible and accessible. Modifications to the tool become infeasible due to convoluted and unreadable code-base that is usually also highly-dependent on external libraries. 

Nyx, on the other hand, aims at different targets. It serves as a transparent, easy-to-understand platform for rapid prototyping and testing of algorithms and approaches. Nyx itself has a very minimalistic structure, only containing 4 base classes (main, parser, planner, and simulator), 8 object classes (e.g., action, state, trace), and 3 compiler classes (for efficient parsing and evaluating of PDDL+). Finally, our planner has two auxiliary template classes for implementing novel heuristic methods and external functions/semantic attachments. Nyx has no complicated or obscure dependencies, it relies only on the very basic concepts in Python. 

Most PDDL+ planners are tailor-built for a specific class of problems and thus only support a subset of PDDL+ features and algorithms. Indeed, there are few planners which are capable of adjusting their planning algorithms and parameters beyond a few simple options. Furthermore, established PDDL+ planners are rigid in their structure and cannot be easily modified to accommodate new classes of planning problems or features. Thus, problems intersecting multiple classes or requiring support for the full set of PDDL+ features have a very limited choice in planning solvers with such coverage. Effectively, each class of PDDL+ problems might require a separate planner suited to that particular type of scenarios, defeating the notion of general-purpose PDDL+ planning.

To counter these shortcomings and ensure that Nyx can deal with a variety of problem types, we equipped Nyx with a range of generic strategies and operating options as a foundation for the planner. However, the built-in functionality of Nyx serves as the basis for further extensions. 
Nyx is specifically designed to adapt to novel classes of problems by allowing semantic and syntactic extensions, and rapid implementations of novel heuristics, external functions, and other features. Nyx facilitates extensibility and straightforward integration of novel features, crucial aspects for attempting to tackle emerging real-world problems and domains whose scope spans beyond PDDL+. 

\subsection{Planning Paradigms}

Simplicity is the main focus in Nyx. The planner relies on fundamental approaches for representing time, states, change, and happenings.
\smallskip

\noindent\textbf{Planning via Discretization} Nyx adheres to the \textit{planning-via-discretization} paradigm for handling the temporal and continuous behavior of hybrid dynamical systems. Nyx approximates a model's continuous system dynamics using a uniform time step ($\Delta t$) and step-functions. Nyx also specifies number precision for representing numeric state variables (which can be specified by the user). 
Nyx follows standard discretization-based planning approach of introducing a special \textit{time-passing} action ($a_{tp}$) that, during search, the planner can choose to apply alongside actions defined in the domain. \textit{Time-passing} allows the planner to advance time during search and apply the effects of events and processes, over a duration of the time step $\Delta t$. \textit{Planning-via-discretization}, while a straightforward approach, enables Nyx to reason with all types of system dynamics, including non-linear behavior. Plans generated by Nyx can be validated (with VAL \cite{howey2004val}) to complete the \textit{Discretize \& Validate} cycle \cite{della2009upmurphi}. \\

\defcitealias{batteryjair}{Fox et al. 2012}
More formally, Nyx converts the planning model into a \textbf{Discretized Finite State Temporal System (DFSTS)} $\mathcal{S} {=} (S, \mathcal{A}, E, P, \mathcal{D}, F)$, where $S$ is a finite set of states, $\mathcal{A}$ is a finite set of actions (including the special time-passing action $a_{tp} {\in} \mathcal{A}$), and $\mathcal{D}$ is a finite set of durations. $F {:} S {\times} \mathcal{A} {\times} \mathcal{D} {\rightarrow} S$ is the transition function. $E$ and $P$ are the finite sets of events and processes, respectively. The formalisms in this work are adapted from \citepalias{batteryjair}. \\

A \textbf{state} $s {\in} S$ is a tuple $s{=}(p(s), v(s), t(s))$, where $p(s)$ is a finite set of propositions, $v(s)$ is a finite set of continuous variables, and $t(s) {\in} \mathbb{R}$ is the time at which state $s$ occurs (i.e., elapsed planning time from the initial state, $t(s_0){=}0$). The transition function $F$ yields a successor state $s' {\in} S$ by executing action $a$ in state $s$, i.e., $F(s,a,d(a)) {=} s'$ where $d(a) {\in} \mathbb{R}_{\geq 0}$ denotes the duration of $a$ and $t(s'){=}t(s)+d(a)$.  \\

An \textbf{action} $a {\in} \mathcal{A}$ is a tuple $a {=} (pre(a), eff(a))$, where $pre(a)$ is a set of preconditions (both propositional and numeric), and $eff(a)$ is a set of effects over state variables $p(s)$ and $v(s)$. An action $a$ is \textit{applicable} in state $s$ iff its preconditions $pre(a)$ are satisfied, i.e., $s {\models} pre(a)$. 
\smallskip

\textbf{Events} and \textbf{processes} are analogous to actions, i.e., $e {=} (pre(e), eff(e)) {\in} E$ and $p {=} (pre(p), eff(p)) {\in} P$, respectively. Thus, their effects can technically be applied via the transition function $F$ from DFSTS, i.e., $F(s,e,d(e)) {\rightarrow} s'$ and $F(s,p,d(p)) {\rightarrow} s'$. In practice, however, events and processes are encapsulated in and executed by the special \textit{time-passing} action $a_{tp}$. The time-passing action has no preconditions itself $pre(a_{tp}) = \emptyset$. On the other hand, the effects of $a_{tp}$ are the union set of effects of all events and processes whose preconditions are satisfied in state $s$: $eff(a_{tp}) {=} \bigcup_{ep {\in} EP} eff(ep) $ where $s {\models} pre(ep)$ and $EP {=} E {\cup} P$.  
\smallskip

A \textbf{DFSTS-based planning problem} is a tuple $\mathcal{P} {=} (\mathcal{S}, s_0, G, \Delta t, T)$, where $\mathcal{S}$ is an DFSTS, $s_0 {\in} S$ is the initial state, $G \subseteq S$ is a finite set of goal states, $\Delta t {\in} \mathbb{R}$ is the discrete time step, and $T$ is a finite temporal horizon which bounds the state space. 
All states (including goal states) must heed the bound imposed by the finite temporal horizon $T$, i.e., $\forall s {\in} S : t(s) \leq T$.  
\smallskip

A \textbf{trajectory} $\pi$ in DFSTS $\mathcal{S}$ is a sequence of states, actions, and action durations, which ends with a state: $\pi = s_i, a_i, d(a_i), s_{i+1}, a_{i+1}, d(a_{i+1}), ..., s_n$, where $i {\in} \mathbb{Z}_{\geq 0}$, $n {\in} \mathbb{Z}_{\geq 1}$, $s_i {\in} S$, $a_i {\in} \mathcal{A}$, and $d(a_i) {\in} \mathcal{D}$. A trajectory $\pi^*$ is a solution to a planning problem $\mathcal{P}$ if it begins in the initial state $s_0$ and ends in a goals state $s_n {\in} G$. Furthermore, for each step $i$ in the trajectory, the transition function $F$ yields the following state, i.e., $F(s_i, a_i, d(a_i)) \rightarrow s_{i+1}$.
\smallskip

Nyx dictates a set of additional assumptions about planning via discretization. All actions are instantaneous (i.e., have duration $d(a)=0$), except \textit{time-passing} $a_{tp}$ whose duration is equal to the discrete time-step $\Delta t$, i.e., $d(a_{tp}) = \Delta t$. Thus, in practice, the finite set of durations is $\mathcal{D} = \{0, \Delta t\}$. In the discretized state space, states only occur at time-points which are multiples of time step $\Delta t$, i.e., $\forall s {\in} S : t(s) = \Delta t * n$, where $n = \mathbb{Z}_{\geq 0}$ is a non-negative integer. Thus, the preconditions of events and processes are also checked at $\Delta t$ intervals. Finally, the effects of continuous processes are applied by the transition function F, each time over the uniform interval $\Delta t$. 
Nyx neglects the support for durative actions which can be seen purely as syntactic convenience. Instead they can be equivalently compiled according to the start-process-stop model \cite{fox2006modelling} where instantaneous actions and/or discrete events can trigger and terminate continuous change.

Finally, setting the time discretization $\Delta t = 0$ nullifies the continuous aspect of the model and removes the need for the time-passing action $a_{tp}$. In Nyx, this is done automatically, by simply excluding \verb|:time| from the PDDL domain requirements. However, Nyx allows events to be triggered in non-temporal domains, since they only apply discrete change themselves. In the 

\smallskip

\noindent\textbf{Grounding} The initial parsing of the PDDL+ domain and problem files explicitly grounds the initial state. All states in Nyx are currently fully grounded. Similarly, all actions, events, and happenings are fully grounded after the PDDL model is parsed. 
\smallskip

\noindent\textbf{Expression Compilation} To bridge the performance gap between Nyx and other planners written in compiled languages, Nyx features expression compilation. For every effect and precondition PDDL expressions in a grounded domain, Nyx translates them to equivalent Python bytecode at runtime. This compilation step is transparent to users and is performed in a just-in-time (JIT) fashion as expressions are evaluated for the first time. Subsequent evaluations of a compiled expression will use JITed Python code instead of interpreted PDDL code for increased efficiency.
\smallskip

\subsection{Search Algorithms and Heuristics}

Similarly to other discretization-based PDDL planners, Nyx converts the discretized PDDL+ problem into a graph search task, where planning states are represented by nodes and actions by edges. Nyx traverses the state space using forward-chaining search algorithms. Currently, the planner is equipped with four basic search algorithms: Breadth-First Search (BFS), Depth-First Search (DFS), Greedy Best-First Search (GBFS), A*\footnote{Currently, the look-back value $g(s)$ of the A* algorithm is implemented to track the number of actions back to the initial state $s_0$. This will be updated in subsequent version of Nyx.}.

By default, breadth-first search is enabled. The user can specify their preferred search algorithm to use by adding a command-line argument when running Nyx. The algorithms change the ordering and queuing procedures of the open list of visited states. Using A* and GBFS requires defining a heuristic function which evaluates each generated successor state. Nyx has a special class which stores the implementations of heuristic functions. Invoking a specified heuristic function is done with command-line parameters. Multiple heuristics can be implemented at the same time, the user only needs to specify the index of the heuristic function they intend to use for a given problem. The heuristic function takes in a planning state object as input and returns a numeric value, i.e., the heuristic estimate for the given state. The heuristic value is then used to enqueue the generated state into the priority open list of explored states. The open list's enqueue mechanism is determined by the active search algorithm, and the states are ordered with respect to the heuristic estimate. 

The planner's search process can be constrained by the user to limit the already vast state space and/or to focus the search effort to a particular subsection. The user can specify the temporal horizon (i.e., states beyond a certain time-point will not be expanded), depth limit (i.e., strict cap on plan length), and planner timeout (limited runtime). 

\subsubsection{Plan Metrics}

Temporal planning is undecidable, thus most planners reasoning with PDDL2.1 and beyond are usually only concerned with finding a feasible solution to a given problem. Due to this obstacle, plan quality has been pushed to the background when working with more advanced domains written in more expressive paradigms such as PDDL2.1 or PDDL+. Plan metrics measure the quality of a solution based on a pre-specified function in the problem file. However, plan metrics have been largely neglected in temporal planning and beyond.

\begin{figure}
\begin{center}
    \fontsize{8pt}{10pt}\selectfont
\begin{verbatim}
(:metric minimize
    (total-time))
(:metric maximize
    (* (fuel_remaining) (total_reward)))
(:metric minimize
    (total-actions))
(:metric minimize
    (* (- (revenue) (cost)) inflation))
\end{verbatim}
\caption{Examples of PDDL plan metrics.}
\label{fig:plan_metrics}
\end{center}
\vspace{-20pt}
\end{figure}

Nyx allows for the parsing of a plan metric defined as a mathematical expression in the PDDL problem file. The metric definition follows the syntax of PDDL+ and requires a function that returns a numeric value. The planning objective is to find solutions with optimized metric (either minimize or maximize). There are two special cases for plan metrics which concern important aspects of the solution but cannot be defined with respect to state variable values. Instead, these metrics focus on finding plans with optimized with respect to makespan (``total-time") and number of plan actions (``total-actions"). If no metric is specified, the planner will minimize plan makespan. 

\subsubsection{Anytime Search}

Nyx is equipped with anytime search algorithm~\cite{hansen1997anytime} for improving the quality of solutions as a trade-off for search time. After the first valid goal state has been encountered, the anytime algorithm exploits prior search progress and continues to explore the state space in search of improved solutions. The anytime approach uses plan metrics which quantify solution quality. Found plans are ranked with respect to plan metrics. After the allotted search time has been exhausted, the planner returns the ranked list of solutions. 

For practicality reasons, the number of solutions concurrently held in memory is limited by the user who can specify the maximum size of the list containing the top plans. Similarly, the user specifies the planning timeout limit. When the limit is reached, Nyx returns the encountered plans. 

Nyx does not have a separate anytime search algorithm as such. Instead, the user can select from the already implemented algorithms (BFS, DFS, GBFS, A*) or implement their own search algorithm. The selected algorithm continues to explore the state space instead of terminating at the first encountered goal state (default timeout is 30 minutes).

\subsection{Nyx Planning Algorithm}

The main planning approach of Nyx is summarized in Algorithm \ref{alg:nyx}. The presented algorithm is a skeleton which can be modified by the user to include various operations, such as an additional event check (at the very beginning of the time passing block). The algorithm returns a set of goal states (\textit{reachedGoals}) from which a trajectory can be extracted via backtracking using \textit{predecessor($s$)} and \textit{achiever($s$)} functions which return the predecessor state and achieving action of any state $s$. 

\setlength{\textfloatsep}{0pt}
\begin{algorithm}[t!]
\SetKwInOut{Input}{Input}\SetKwInOut{Output}{Output}
\Input{$\mathcal{A}_g$, $E_g$, $P_g$ - list of grounded actions, events, and processes, respectively.}
\Input{requirements - list of domain requirements.}
\Input{$s_0$ - initial state.}
\Input{$\Delta t$ - time discretization quantum.}
\Output{reachedGoals - list of reached goal states.}
OPEN$\gets\{s_0\}$\\ 
\While{OPEN}{
    $s\gets$ pop(OPEN)\\

    \If{$s \in G$}{
        enqueue(reachedGoals, $s$, metric($s$))\\
        \If{not anytime}{
            return reachedGoals
        }
    }
    
    $s' \gets$ copy($s$)\\
    predecessor($s'$) $\gets s$\\
    
    \ForEach{$a \in$ applicable(actions, $s$)}{
        \eIf{$a$ is time-passing}{

            \If{``:semantic-attachment" $\in$ requirements}{
                $s' \gets$ semanticAttachment($s'$)
            }
            
            \ForEach{$p \in$ applicable(processes, $s'$)}{
                $s' \gets $ $F$($s$,$p$, $\Delta t$)
            }
            \ForEach{$e \in$ applicable(events, $s'$)}{
                $s' \gets $ $F$($s$,$e$,$0$)
            }
        }
        {
            $s' \gets$ $F$($s$,$a$, $0$)\\
            
            \If{``:time" $\notin$ requirements}{
                \ForEach{$e \in$ applicable(events, $s'$)}{
                    $s' \gets $ $F$($s$,$e$, $0$)\\
                }
            }
        }
        achiever($s'$) $\gets a$\\
        \If{isValid($s'$)}{
            insert(visited, $s$)\\
            enqueue(OPEN, $s'$, $h(s')$)\\
        }
    }
    \If{runtime limit reached}{
        \textbf{break}
    }
}
\Return reachedGoals\\
\caption{Nyx Planning Algorithm}
\label{alg:nyx}
\end{algorithm}




\subsection{Precondition Tree}
Internally, Nyx reasons with grounded representations of states, actions, events, and processes. Unfortunately, this approach can cause an explosion in the number of happenings which need to be repeatedly checked for applicability at every step during search. Checking a list of preconditions of all actions, events, and processes in a linear fashion is often unfeasible. To improve the efficiency of the planner, this paper introduces \textbf{precondition tree}, a novel approach to checking happening applicability inspired by the concept of the trie~\cite{fredkin1960trie}. Structuring the precondition checking as a tree traversal task allows for efficient pruning of actions which contain falsified preconditions. A single falsified precondition can prune multiple actions from being unnecessarily checked for applicability.

A \textbf{precondition node} of a precondition tree is a tuple\\ $pn {=} (expr(pn), \mathcal{A}_{expr}(pn), C(pn))$, where $\forall a {\in} \mathcal{A}_{expr}(pn) {:} expr(pn) {\in} pre(a)$ is a propositional or numeric precondition expression, a finite set of actions $\mathcal{A}_{expr}(pn) {\subseteq} \mathcal{A}$ which all contain the $expr(pn)$ precondition, i.e., $\forall a {\in} \mathcal{A}_{expr}(pn) : expr(pn) {\in} pre(a)$. $C(pn)$ is a finite set of child precondition nodes. 

Note, that for each precondition node $pn$, each action in $\mathcal{A}_{expr}(pn)$ also contains all preconditions from its parent nodes. Thus, for each action $a {\in} \mathcal{A}$, its set of preconditions $pre(a)$ is represented as a trajectory (a sequence of linked precondition nodes) in the precondition tree, where the final node of the trajectory contains the action $a$. 

More formally, a \textbf{precondition trajectory} is a sequence of precondition nodes $\tau {=} pn_i, pn_{i+1}, ..., pn_n$, where $i {\in} \mathbb{Z}_{\geq 0}$ and $n {\in} \mathbb{Z}_{\geq 1}$. Each precondition node $pn$ contains the following node inside its set of child nodes, $\forall pn_k {\in} \tau {:} pn_{k} {\in} C(pn_{k-1})$ where $k {\in} \mathbb{Z}_{\geq 1}$. The final node $pn_n$ of any trajectory $\tau$ must contain a non-empty set of actions $\mathcal{A}_{expr}(pn_n) \neq \emptyset$. The set of preconditions $pre(a)$ for any action $a {\in} \mathcal{A}_{expr}(pn_k)$ contained inside some precondition node $pn_k {\in} \tau$ must also contain the preconditions contained by all of its predecessor nodes in the precondition trajectory, i.e., $\forall a {\in} \mathcal{A}_{expr}(pn_k) {:} pre(a) = \bigcup^{k}_{i = 0} expr(pn_i)$. 

In a precondition tree, multiple precondition trajectories can partially overlap. In fact, a precondition trajectory of some action $a_k {\in} \mathcal{A}$ may contain a whole precondition trajectory of another action $a_j {\in} \mathcal{A}$ iff $pre(a_j) \subset pre(a_k)$. If $pre(a_j) = pre(a_k)$, they would be part of the same trajectory $\tau$ with the final precondition node $pn_n$ containing both actions, i.e., $a_j, a_k {\in} \mathcal{A}_{expr}(pn_n)$

Given a grounded state $s {\in} S$, the mechanism traverses the precondition tree, starting at the root, evaluating the precondition expression $expr(pn_i)$ in each expanded node $pn_i$. If the grounded precondition is satisfied in the given state, all actions contained by that precondition node are deemed applicable, i.e., $\forall a {\in} \mathcal{A}_{expr}(pn_i) : s \models expr(pn_i) {\implies} s {\models} pre(a)$. Furthermore, if $s \models expr(pn_i)$ then all child nodes $C(pn_i)$ will be expanded and evaluated. Conversely, if precondition $expr(pn_i)$ is falsified in $s$, i.e., $s \not\models expr(pn_i)$, the entire branch containing precondition node $pn_i$ and all of its successors is pruned away. The precondition tree traversal is shown in fig. \ref{alg:tree_traversal}. 

\begin{algorithm}[t!]
\SetKwInOut{Input}{Input}\SetKwInOut{Output}{Output}
\Input{$s \in S$ - a planning state.}
\Input{$pn_{rt} = (expr(pn_{rt}), \mathcal{A}_{expr}(pn_{rt}), C(pn_{rt})$ - root node of a grounded precondition tree.}
\Output{APPLICABLE - list of applicable actions.}
OPEN$\gets\{pn_{root}\}$\\ 
APPLICABLE $\gets \emptyset$\\
\While{OPEN}{
    $pn \gets$ pop(OPEN)\\

    \If{$s \models expr(pn)$}{
        \ForEach{$a \in \mathcal{A}_{expr}(pn)$}{
            APPLICABLE.append($a$)\\
        }
        \ForEach{$pn_c \in C(pn)$}{
            OPEN.append($pn_c$)\\
        }
    }
}
\Return APPLICABLE\\
\caption{Precondition tree traversal to find applicable grounded actions.}
\label{alg:tree_traversal}
\end{algorithm}

A precondition tree is generated by iteratively inserting a precondition node $pn$ for each precondition $expr {\in} pre(a)$ for each action $a {\in} \mathcal{A}$ in the grounded planning problem $\mathcal{P}$. Algorithm for generating a precondition tree is shown in \ref{alg:tree_generation}.

\begin{algorithm}[t!]
\SetKwInOut{Input}{Input}\SetKwInOut{Output}{Output}
\Input{$\mathcal{A}$ - a set of grounded actions\\ $\forall a \in \mathcal{A} : a = (pre(a), eff(a))$.}
\Input{$pn_{rt} = (expr(pn_{rt})=\emptyset, \mathcal{A}_{expr}(pn_{rt})=\emptyset, C(pn_{rt})=\emptyset)$ - an empty root precondition node.}
\Output{$pn_{rt}$ - a root precondition node (containing the precondition tree).}


\ForEach{$a \in \mathcal{A}$}{
    $pn \gets pn_{rt}$\\
    \ForEach{$expr \in pre(a)$}{
        
        $pn_{match} \gets \emptyset$\\

        \eIf{$C(pn) = \emptyset$}{
            $pn_{match} = (expr, \emptyset, \emptyset)$\\
            $C(pn)$.append($pn_{match}$)\\
        }
        {
            \ForEach{$pn_c \in C(pn)$}{
                \If{$expr(pn) = expr(pn_c)$}{
                    $pn_{match} \gets pn_c$\\
                    \textbf{break}
                }
            }
        }
        $pn \gets pn_{match}$
    }
}
\Return $pn_{rt}$\\
\caption{Precondition Tree Generation}
\label{alg:tree_generation}
\end{algorithm}

Since all happening types (i.e., actions $\mathcal{A}$, events $E$, and processes $P$) have identical structure, the precondition tree approach is applicable to all happenings. Algorithms \ref{alg:tree_generation} and \ref{alg:tree_traversal} can be adapted by replacing the set of grounded actions $\mathcal{A}$, with the set of events $E$ or processes $P$. In practice, Nyx generates a precondition tree for each type of happenings.

\defcitealias{howey2004val}{Howey et al. 2004}
From a design perspective, the precondition tree is a more efficient approach instead of sequentially checking lists of preconditions to determine applicability of each grounded happening. In practice, however, the efficiency of the precondition tree is affected by the planning domains and their characteristics. 
Many domains have small branching factors, relatively few grounded happenings of each type, and simple precondition expressions. Under those circumstances, building the precondition tree can add unnecessary overhead. As a result, the precondition tree is an optional feature in Nyx that can be activated via a command-line flag when needed. Table \ref{tab:precondition_tree_results} presents the impact of the precondition tree on search performance, measured w.r.t. the average number of explored states per second over the first 10000 states. As is evident from the data, the precondition tree approach more than doubled the state space exploration rate for the Angry Birds domain \cite{piotrowski2021playing} where 1413 grounded events need to be checked at each time tick (checking for collisions, explosions, and other phenomena between birds, pigs, blocks, and platforms). It also has the highest average number of preconditions for all happening types and the vast majority of preconditions are complex numeric expressions which are particularly time-consuming to evaluate. However, the event-driven Angry Birds model is an outlier in its complexity compared to other benchmark PDDL+ models (\citet{fox2006modelling, mcdermott2003reasoning, stern2022model}; \citetalias{howey2004val}).

\begin{table*}[]
\small
\resizebox{\textwidth}{!}{%
\begin{tabular}{|c|c|c|c|ccl|}
\hline
\multirow{2}{*}{\textbf{Domains}} &
  \multirow{2}{*}{\textbf{\begin{tabular}[c]{@{}c@{}}\# actions\\ (avg \# preconditions)\end{tabular}}} &
  \multirow{2}{*}{\textbf{\begin{tabular}[c]{@{}c@{}}\# events\\ (avg \# preconditions)\end{tabular}}} &
  \multirow{2}{*}{\textbf{\begin{tabular}[c]{@{}c@{}}\# processes\\ (avg \# preconditions)\end{tabular}}} &
  \multicolumn{3}{c|}{\textbf{exploration rate (nodes/sec)}} \\ \cline{5-7} 
                &          &            &           & \multicolumn{1}{c|}{\textbf{with PT}} & \multicolumn{1}{c|}{\textbf{without PT}} & \textbf{difference} \\ \hline
\textbf{Car}             & 4  (2.33) & 1  (3)      & 2  (1.5)   & \multicolumn{1}{c|}{44971}         & \multicolumn{1}{c|}{54489}            & -17.5\%             \\ \hline
\textbf{Sleeping Beauty} & 4  (1.67) & 6  (2.5)    & 2  (1.5)   & \multicolumn{1}{c|}{41037}         & \multicolumn{1}{c|}{54132}            & -24.2\%             \\ \hline
\textbf{Vending Machine} & 3  (2)    & 3  (3)      & 1  (1)     & \multicolumn{1}{c|}{32932}         & \multicolumn{1}{c|}{37260}            & -11.6\%             \\ \hline
\textbf{Convoys}         & 129  (2)  & 0  (0)      & 65  (1.98) & \multicolumn{1}{c|}{6305}          & \multicolumn{1}{c|}{7841}             & -19.6\%             \\ \hline
\textbf{Cartpole}        & 3  (4)    & 4  (2)      & 1  (2)     & \multicolumn{1}{c|}{10846}         & \multicolumn{1}{c|}{11269}            & -3.7\%              \\ \hline
\textbf{Angry Birds}     & 9  (5.5)  & 1413  (7.1) & 5  (3.6)   & \multicolumn{1}{c|}{924}           & \multicolumn{1}{c|}{422}              & +118.8\%            \\ \hline
\end{tabular}%
}
\caption{Comparison of Nyx's exploration rate (w.r.t. expanded nodes) with and without using the precondition tree (PT) approach for various PDDL+ domains. The total number of grounded actions, grounded events, and grounded processes is presented per domain. The average number of preconditions per each happening type is show in brackets.}
\label{tab:precondition_tree_results}
\vspace{-10pt}
\end{table*}

Finally, like any approach for checking happening applicability, the precondition tree is sensitive to the ordering of preconditions. For uniformity, all sets of preconditions are sorted in lexicographical order. Though, out of scope for this paper, subsequent work will conduct an analysis of node ordering in the precondition tree to optimize its efficiency.

\section{Nyx Expressiveness Extensions}

PDDL+ is one of the most expressive planning modeling languages in use today. It is specifically designed as a representation for hybrid systems with mixed discrete and continuous dynamics. Hybrid systems are omnipresent in the real world. In fact, most realistic systems exhibit both discrete and continuous behavior. Furthermore, PDDL+ is also able to express exogenous activity in the form of processes and events (i.e., the environment's actions). Deployed intelligent agents are required to interact with real-world phenomena, making PDDL+ well-suited for modeling realistic scenarios. 

However, PDDL+ models are still severely limited by the classes of mathematical expressions they can define. Currently, PDDL+ (and other numeric versions of PDDL) are limited to basic arithmetic operations, i.e., addition, subtraction, multiplication, and division. Thus, any significantly advanced system dynamics, that cannot be easily defined using the aforementioned operations, must be simplified or approximated. This is at odds with the real-world planning applications that require model accuracy and greater expressiveness. Indeed, even quite basic mathematical operations such as roots, absolute value, or trigonometric functions are very difficult, and often impossible, to accurately encode in a PDDL planning domain. 

Some approximations are sufficiently accurate to exploit (e.g., Bhaskara's trigonometric approximations) but this often comes at a cost of significant reduction in model clarity, readability, and/or conciseness. Figure~\ref{fig:bhaskara_pddl_approximation} shows the PDDL expression required to approximate $\sin{\theta^\circ}$ and $\cos{\theta^\circ}$. Modeling complex systems using bloated and overly complicated approximations for basic mathematics is cumbersome and time-consuming. Additionally, such practices are prone to introducing errors into the models. Most importantly, having to resort to using much simplified dynamics or complicated approximations may discourage users from considering AI planning approaches altogether. 

Nyx was specifically built with a focus on extensibility and adaptability. It was also designed to overcome the severe limitations imposed by a rigid code architecture, common in most other PDDL+ planners. Nyx facilitates straightforward extensions for defining advanced system dynamics beyond the current arithmetic confines of default PDDL+.

\begin{figure*}[h]
\begin{center}
\begin{verbatim}

sin_theta= (/ (* (* 4 (theta)) (- 180 (theta))) (- 40500 (* (theta) (- 180 (theta)))))
cos_theta= (/ (- 32400 (* 4 (* (theta) (theta)))) (+ 32400 (* (theta) (theta))))
\end{verbatim}
\caption{PDDL-style implementation of sin($\theta^\circ$) and cos($\theta^\circ$) using Bhaskara's approximation.}
\label{fig:bhaskara_pddl_approximation}
\end{center}
\vspace{-10pt}
\end{figure*}

\subsubsection{Domain Language Extensions}

Currently, realistic system models require significantly more advanced dynamics than can be expressed using basic arithmetic. Yet, PDDL+ does only support such fundamental operations. Poor support for mathematical expressions is one of the major limiting factors of planning models.

Nyx supports extending the expressive power of PDDL+ by integrating new mathematical expressions and operations. The planner's parser is written explicitly in Python (rather than using libraries such as flex), ensuring transparency when processing PDDL models. Nyx facilitates seamless integration of the new expressions into the planner in a straightforward manner. The user is only required to add a new entry to an existing list of mathematical symbols/function names, and then define how the parser will evaluate the expression. In practice, adding a new expression requires only about 2 lines of code in one file (one specifying the new symbol and arity of the expression, the other defining how to compute the expression in Python).

Nyx allows the PDDL files to contain the $^\wedge$ symbol representing a power expression in the form of \verb+(^ (x) (y))+, where x is the base number and y is the exponent. This power expression also accepts fractional exponents which allows for representation of roots. Figure~\ref{fig:absolute_value_pddl} shows an example usage of the $^\wedge$ operator to compute the absolute value of a variable $|z|$, as well as the magnitude of a vector $|\vec{v}|$. 

\begin{figure}
 \fontsize{8pt}{10pt}\selectfont
\begin{center}
$\displaystyle |x| = \sqrt{x^2}$
\begin{verbatim}
    (assign x_abs (^ (^ (z) 2) 0.5))

\end{verbatim}

$\displaystyle |\vec{v}| = \sqrt{(x_2 - x_1)^2 + (y_2 - y_1)^2}$
\begingroup
    \fontsize{8pt}{10pt}\selectfont
\begin{verbatim}
(assign v_mag 
  (^ (+ (^ (-(x2)(x1)) 2) (^ (-(y2)(y1)) 2)) 0.5))
\end{verbatim}
\endgroup

\caption{Absolute value of some variable $z$ and the magnitude of some vector $\vec{v}$, using Nyx's added power operator $^\wedge$.}
\label{fig:absolute_value_pddl}
\end{center}
\end{figure}

PDDL+, while expressive, lacks support for important mathematical operations that are required to model realistic systems. Thus, flexibility and ease of implementing extensions of the modeling language can be crucial in tailoring Nyx for solving any real-world problems. Virtually any type of operator can be seamlessly integrated into Nyx, including specialized expressions which require external libraries/packages (provided they are imported in the parser). Finally, while the arithmetic operators in PDDL are binary, newly added expressions in Nyx can have unrestricted operand cardinality. 

\subsubsection{Semantic Attachments}

Real-world systems can be extraordinarily complex. Unfortunately, some realistic system dynamics cannot be defined as part of a planning domain even exploiting Nyx's domain language extensions described in the previous section. Such cases represent a significant loss of scientifically interesting scenarios to the planning community. Similarly, the target application stakeholders may dismiss automated planning as not sufficiently intricate or expressive and, instead, turn to other approaches. In many cases, modeling a real-world system as a planning domain might prove unfeasible due to a small aspect or feature of the system which cannot be encoded using PDDL+ syntax. 

To overcome scenarios where a piece of the system dynamics is beyond the scope of expressiveness of PDDL+, Nyx is equipped to accommodate semantic attachments~\cite{dornhege2009semantic, semantic_attachements}. Semantic attachments are external functions to which the planner delegates the computation of some of the system's dynamics. Via semantic attachments, Nyx enables the integration of advanced methods and libraries which cannot be exploited otherwise. Furthermore, it facilitates a straightforward manner of integrating external functions in virtually any form, including pure Python code, trained surrogate models, executable simulators, to name a few. 

Nyx models a semantic attachment as a Python function that takes a planning state as input, updates a subset of state variables' values, and returns a modified state. In the planner, the feature is activated by simply adding \verb|:semantic-attachment| to the PDDL domain requirements. By default, the semantic attachment is invoked inside the time-passing action in between applying change from the triggered events and active processes. However, this ordering can be modified in a trivial manner to call the external function at the appropriate stage suited for the given application.

\begin{figure}[h]
\begin{center}
\includegraphics[width=0.99\columnwidth]{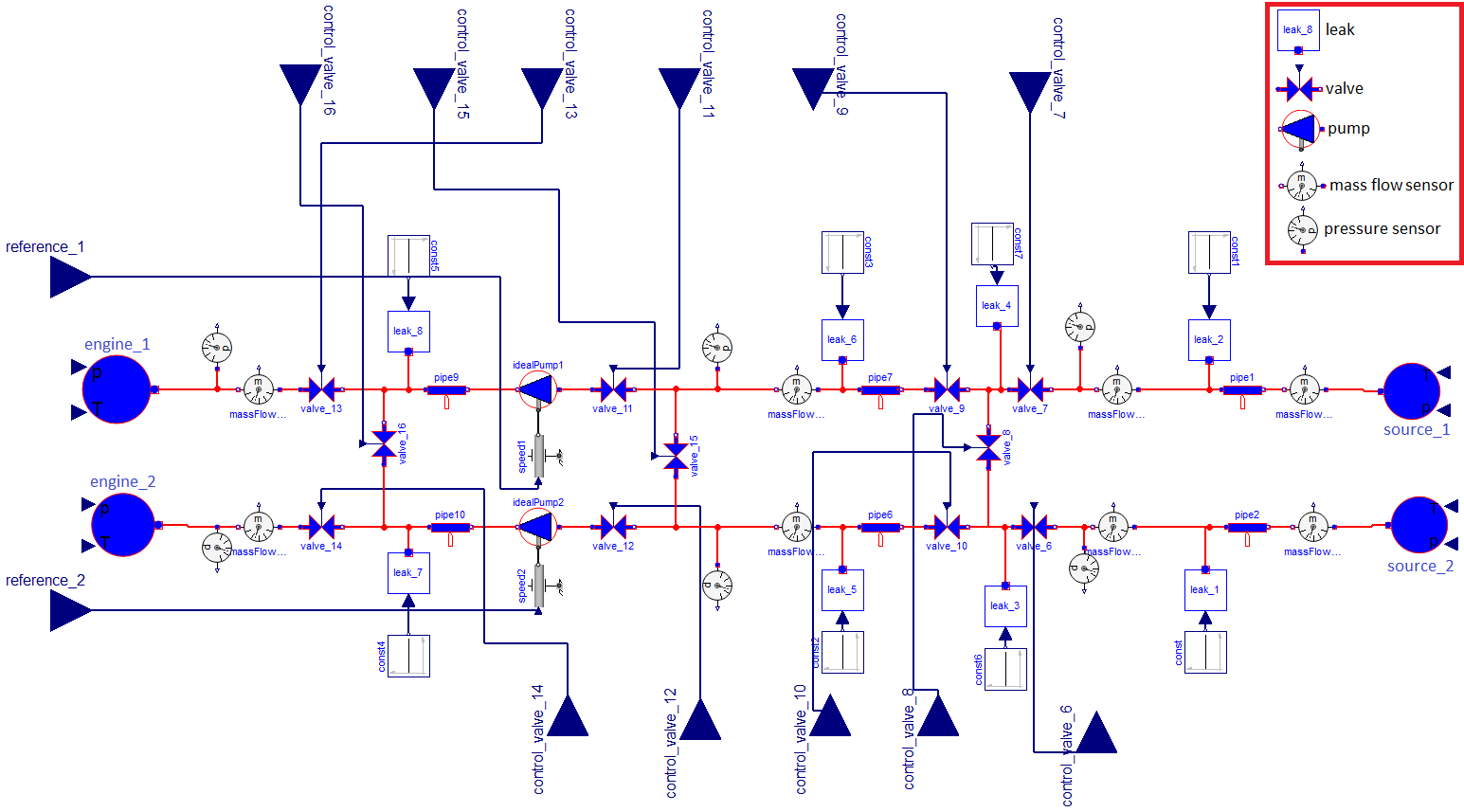}
\caption{Fuel system Modelica model simulated inside Nyx's semantic attachment as an FMU.}
\label{fig:fuel_system}
\end{center}
\vspace{-10pt}
\end{figure}

Nyx's semantic attachment approach has already been successfully employed in several real-world applications. The most notable example is automatic reconfiguration of a ship-board fuel system~\cite{matei2022system} which integrated a Functional Mock-up Unit (FMU) as the semantic attachment. FMU~\cite{blochwitz2011functional} is a standardized executable used to simulate models when developing complex cyber-physical systems. It is an established standard for model exchange and independent simulation in the field of Engineering Design and can be exported from various system design tools such as Simulink, Modelica, or Dymola. In the fuel system scenario, Nyx was tasked with system reconfiguration to isolate or mitigate the effects of pipe leaks by adjusting the positions of valves and speeds of pumps controlling the fuel flow. The computation of fluid dynamics was delegated to the FMU via the semantic attachment which updated the values of state variables representing fuel mass-flow rates at the engines and tanks. Nyx was successful in rerouting the fuel to avoid the leaky pipe and minimizing fuel loss while simultaneously ensuring the engines are fed sufficient levels of the propellant to complete the ship's mission. The Modelica model of the fuel system used to generate the FMU is shown in fig.~\ref{fig:fuel_system}.

Nyx was also used in other real-world scenarios where semantic attachments were set up to: compute dynamic fuel consumption rate via SciPy-based interpolation based on vehicle velocity and engine torque level; and compute power generation from solar panel arrays, batteries, and diesel engines with a slipping clutch. The latter simulator was tested in two different formats: FMU-based simulation via PyFMI, and a DNN-based surrogate model via PyTorch. 

\section{Discussion}

\begin{table}[]
\resizebox{\columnwidth}{!}{%
\begin{tabular}{|l|l|l|l|l|l|}
\hline
\multicolumn{1}{|c|}{\textbf{Planner}} &
  \multicolumn{1}{c|}{\textbf{Cartpole}} &
  \multicolumn{1}{c|}{\textbf{\begin{tabular}[c]{@{}c@{}}Angry\\ Birds\end{tabular}}} &
  \multicolumn{1}{c|}{\textbf{\begin{tabular}[c]{@{}c@{}}Sleeping\\ Beauty\end{tabular}}} &
  \multicolumn{1}{c|}{\textbf{\begin{tabular}[c]{@{}c@{}}Vending\\ Machine\end{tabular}}} &
  \multicolumn{1}{c|}{\textbf{Car}} \\ \hline
Nyx      & 12020 & 433   & 50528  & 34930  & 52179 \\ \hline
UPMurphi & 53800 & 334   & 147800 & 108140 & 81900 \\ \hline
ENHSP    & 27990 & 11851 & 39227  & 30434  & 43747 \\ \hline
\end{tabular}%
}
\caption{Representative rates of state space exploration in expanded nodes per second.}
\label{tab:perf_comparison}
\end{table}

Real-world systems are challenging for AI with respect to modeling and solving them. Currently available PDDL+ planners have a steep learning curve and require expert knowledge. Nyx aims to increase the accessibility to AI planning, particularly for realistic feature-rich domains, by focusing on simplicity and adaptability. However, such advantages come at a cost of raw computational performance compared to complex optimized code and low-level languages such as C/C++. For transparency, Table~\ref{tab:perf_comparison} shows comparison with UPMurphi and ENHSP\footnote{These domains are an intersection of models that can be handled by all three planners. Still, Cartpole, Sleeping Beauty, and Vending Machine needed minor refactoring for ENHSP. This is further evidence of difficulties in using Planners.} (both using blind BFS) in representative speed of state space exploration in nodes per second. Crucially, it should be noted that Nyx addresses a different classes of problems from the other planners. Our planner facilitates rapid prototyping of planner extensions and novel heuristics, solving emerging classes of domains which cannot be handled by any established planners, and reasoning with composite PDDL+ models with external functions \& features. 

\section{Conclusion \& Future Work}

This paper presented Nyx, a novel PDDL+ planner for real-world planning problems. Nyx's design facilitates adaptability to tackle novel classes of domains, and accessibility to promote AI Planning as a viable and usable method for solving interesting real-world problems. We also present the precondition tree, a promising new approach to efficiently evaluating preconditions. Furthermore, Nyx introduces features that support reasoning advanced features beyond the scope of PDDL+. Specifically, Nyx allows straightforward implementation of new mathematical expressions to be used in PDDL+ domains, as well as support for semantic attachments for features that cannot be feasibly included in the PDDL model directly. By discussing the fuel system model, we show that Nyx enables AI Planning to reason with real-world problems. In future work, we will continue exploring different configurations of the precondition tree and analyze impact of different orderings of preconditions in the tree. We will also work on improving the user interface and implementation interface of Nyx to further reduce the accessibility barrier. 

\newpage

\bibliography{library}
\end{document}